\documentclass[runningheads]{llncs}

\usepackage{graphicx}
\usepackage{multirow}
\usepackage{marvosym}

\begin{document}

\title{A Survey of Pedophile Attribution Techniques for Online Platforms \thanks{Published in American Journal of Humanities and Social Science, Vol. 33, 2022}
}

\author{Hiba Fallatah \Letter \orcidID{0000-0001-6134-9466} \and
Ching Suen\orcidID{0000-0003-1209-7631} \and
Olga Ormandjieva\orcidID{0000-0001-5641-9765}}

\authorrunning{H. Fallatah et al.}

\institute{Concordia University, Montr\'eal, Qu\'ebec, Canada\\
\email{h\_fall@encs.concordia.ca}}

\maketitle              
\begin{abstract}

Reliance on anonymity in social media has increased its popularity on these platforms among all ages. The availability of public Wi-Fi networks has facilitated a vast variety of online content, including social media applications. Although anonymity and ease of access can be a convenient means of communication for their users, it is difficult to manage and protect its vulnerable users against sexual predators. Using an automated identification system that can attribute predators to their text would make the solution more attainable. In this survey, we provide a review of the methods of pedophile attribution used in social media platforms. We examine the effect of the size of the suspect set and the length of the text on the task of attribution. Moreover, we review the most-used datasets, features, classification techniques and performance measures for attributing sexual predators. We found that few studies have proposed tools to mitigate the risk of online sexual predators, but none of them can provide suspect attribution. Finally, we list several open research problems.

\end{abstract}

\section{Introduction}
In the past decade, communicating through online platforms has facilitated a vast variety of online content and services and enabled us to overcome geographical boundaries. Although these features provide a convenient means of communication for their users, and an ideal alternative to work and learn from a distance, it is difficult to manage and protect its vulnerable users, particularly children, against online predators. Detecting and identifying these predators is challenging \cite{rocha2016authorship}.

Sexual predatory behaviour or “Pedophilia” is defined in \cite{Gavin2013Criminological} as a disorder in which an adult or older adolescent experiences a primary or exclusive sexual attraction to children. From this definition, we can establish that online pedophiles are people who “groom” children, that is, who meet underage victims online, engage in sexually explicit text or video chat with them, and eventually convince the children to meet them in person \cite{villatoro2012two}. According to Statistics Canada Police-Reported crime data: luring a child via computer incidents has increased in 2020 by 244 cases, representing a 15 percent increase compared to 2019 across Canada. This is a 37 percent increase in incidents compared to the previous five-year average\cite{Canadacrime2021,Police-reported2021}.
Anonymity in social media, ease of access and the massive amount of data made detecting and identifying sexual predators manually a difficult task. Using an automated identification system that can identify and attribute predators to their text would make the problem more manageable. An automated approach can help process an extensive amount of data and reveal hidden patterns that would not be noticed otherwise. Thus, it is qualified to be used by law enforcement.
\subsection{Definition}
In \cite{juola2008authorship} the author has defined “authorship attribution” broadly as any attempt to infer the characteristics of the creator of a piece of linguistic data. In other words, the goal of authorship attribution is to identify authors of texts through features derived from the style of their writing; this is called Stylometry \cite{rocha2016authorship}.

In the early 1960s, Mosteller and Wallace worked on a detailed study on the authorship of “ The Federalist Papers” \cite{stamatatos2009survey}. Their work was considered the base of computer-assisted stylometry \cite{koppel2011authorship,mosteller1963inference,mosteller1961notes}. From then to the late 1990s most of the stylometry literature has focused on defining features to quantify writing style \cite{holmes1994authorship,mosteller1963inference}. Since the late 1990s a large amount of text, emails, blogs, and online forms have been generated, in addition to the development of information retrieval and the areas of machine learning and natural language processing, the focus has shifted to evaluate the proposed methods, and compare different methods based on benchmark corpus \cite{stamatatos2009survey}.  We can detect some of the early work that used author attribution techniques as a forensic tool for the available online platforms back then. For example, De Vel et al. in \cite{koppel2011authorship} used it for emails to investigate the learning of authorship categories for the case of both aggregated and multi-topic email documents. Zheng et al. in \cite{li2006fingerprint} developed a generic algorithm-based feature selection model to identify the key features of write print specifically for online forms, and Gray et al. in\cite{gray1998identified} were even able to build a toolkit called IDENTIFIED that can be used for software forensics.

In \cite{koppel2011authorship}, Koppel et al. have stated the two main parameters that affect the performance of authorship attribution analysis: the number of candidate authors, and the size of the training text. Stamatatos et al. in\cite{stamatatos2009survey} added the distribution of the training corpus over the authors (balanced or imbalanced data) as an extra consideration. In the following sections, we will discuss how each of these parameters affects the process of attribution, especially in terms of pedophile attribution. 

In this survey, we will provide an author attribution overview with a focus on the online pedophile's attribution task. A discussion of how the size of the candidate set (i.e. number of suspects) can affect the attribution task is introduced in section two. Also, in section three we will see how short text attribution is different from the attribution of a long one. Section four will list the steps of authorship attribution. The following section will present the most used performance analysis matrices. Then, we will provide a comparison between the different tools for this task. Open research problems in the field will follow, and finally the conclusion in the last section.

\section{Authorship Attribution and Number of Candidates}

According to \cite{juola2008authorship}, in general, there are three author attribution problems found in the literature regarding the number of authors. First, the ” closed class” problem: given a particular sample of text known to be by one of a set of authors, determine which author wrote that sample. Closed class- or closed set as it is referred to in some of the literature presumes that the text that needs to be examined belongs to a limited finite set of authors. This means that all the authors seen at the testing time are also known at training time \cite{rocha2016authorship}. Most of the authorship attribution application and experimental effort is directed towards the closed set task \cite{forstall2011evidence}.

Second, The “open class” problem: given a particular sample of text believed to be by an unknown author, having a set of authors, determine which one wrote the text, if any \cite{juola2008authorship}. Unlike the previous class, the open class attribution author set is unlimited. The text being tested can belong to one of the authors in the list or none of them at all. This is a much harder problem than the closed set, although it is the most occurring in a real-world setting\cite{juola2008authorship,koppel2011authorship,rocha2016authorship}. For one, the classifier has no prior assumption made on the testing data \cite{koppel2011authorship}, i.e. the author set is not exclusive. In addition, in some cases, an author’s style is very similar to that of other authors, as they may have similar education, influence or culture \cite{rocha2016authorship}, so the attribution of the text has a higher chance of being faulty and not accurate. In \cite{koppel2011authorship} authors considered the case when the candidate doesn’t belong to the author set, they found that having fewer candidates makes the open-set problem more difficult and the model could mistakenly attribute the text to one of the candidate authors.

The third attribution problem is “profiling”. In this problem, the goal is to determine any of the properties of the author(s)(e.g., age, gender) of a sample of text \cite{juola2008authorship,stamatatos2009survey}. Profiling helps reduce the large pool of possible authors to a manageable set of candidates without directly identifying the author \cite{neal2017surveying}. 

It is worth mentioning, that in the special case when the attribution problem has a candidate set size equal to one, it is called an authorship verification problem. Authorship verification is typically a binary classification problem that decides if two documents were written by the same author\cite{halvani2016authorship}.  According to the research in [51],  Author verification can be considered as the "fundamental problem" in authorship attribution, thus categorizing it as the fourth attribution problem. 

Considering the number of authors (i.e. number of classes to be learned) when these authors are pedophiles, usually, we are dealing with a verification problem, where we know the identity of the online offender and we are trying to prove his authorship of an anonymous text message. Unfortunately, in most cases, it's also an open-set problem where the set of suspects (i.e. authors) is open and unlimited.

\section{Authorship Attribution and Text Size}

As mentioned earlier, the idea of using statistical pattern recognition and computer linguistic methodologies to perform an automated authorship attribution started decades ago. Since then, the main focus has been on the digital humanities, where the goal is to verify a disputed written text \cite{binongo2003wrote,craig1999authorial,kocher2017simple}. However, with the widespread use of online social media and the massive amount of data that comes with it, researchers have given attention to authorship analysis for this type of data, especially for the forensic domain. For example, the work of Rocha et al. in \cite{rocha2016authorship}, and Stamatatos et al. in \cite{stamatatos2009survey}, and  Zheng et al. in \cite{zheng2006framework} have focused on social media for forensic, including; authorship attribution, verification, and identification. With that attention, researchers have noticed that although the task of authorship attribution analysis can be similar for both long text attribution and short text attribution in some aspects, they have many differences that prevent them from using the same techniques for both types of texts \cite{luyckx2011effect}.

Longer texts like novels, news articles, and books tend to provide a wealth of features that can be computed and extracted and then analyzed to pinpoint its truthful author \cite{rocha2016authorship}. Also, this type of text is usually written in a formal uniform tone, with few or even no spelling or grammatical errors, which makes the preprocessing step less tricky.

On the other hand, we don’t have these privileges with short text messages. In the real world, we only get provided with a few short messages from an anonymous author that we need to analyze in order to extract his or her stylistic features, which is not enough to come up with a near-accurate estimation of the right owner of these messages \cite{luyckx2011effect,neal2017surveying,rocha2016authorship}. In addition, people tend to write informally on social media and sometimes they use unfamiliar phrases, characters, images or abbreviations that need to be considered when we are trying to pre-process the data \cite{ebrahimi2016automatic}.

Authors of \cite{rocha2016authorship} have listed the key considerations for forensic authorship attribution for social media as:
\begin{itemize}
\item The samples provided for the testing set could be limited (i.e. there are not enough samples).
\item The training set that is used to build the prediction model could have some noise, and the
distribution of the samples among authors is not balanced (some authors could have more
samples than others). In other words, there is no guarantee of the quality of the testing samples.
\item The process of the system needs to be well-defined to build efficient algorithms.
\item The error rate used for an algorithm needs to be well determined, as having a high probability of wrong predictions could have legal ramifications.
\item The used algorithm should always be designed with an adversarial scenario in mind (i.e. suspect can avoid automated attribution methods). 
\end{itemize}

Indeed, these considerations also can be applied when the task is attributing the harmful text to online predators. These predators usually target their victims on online platforms where the communication is often delivered through short messages and under a fake identity. In these types of settings, researchers often need to deal with a single short text of the suspect author, so attribution will be less reliable than expected from reported results \cite{luyckx2011effect}.

\section{Steps of Authorship Attribution}

\begin{figure}
\includegraphics[width=\textwidth]{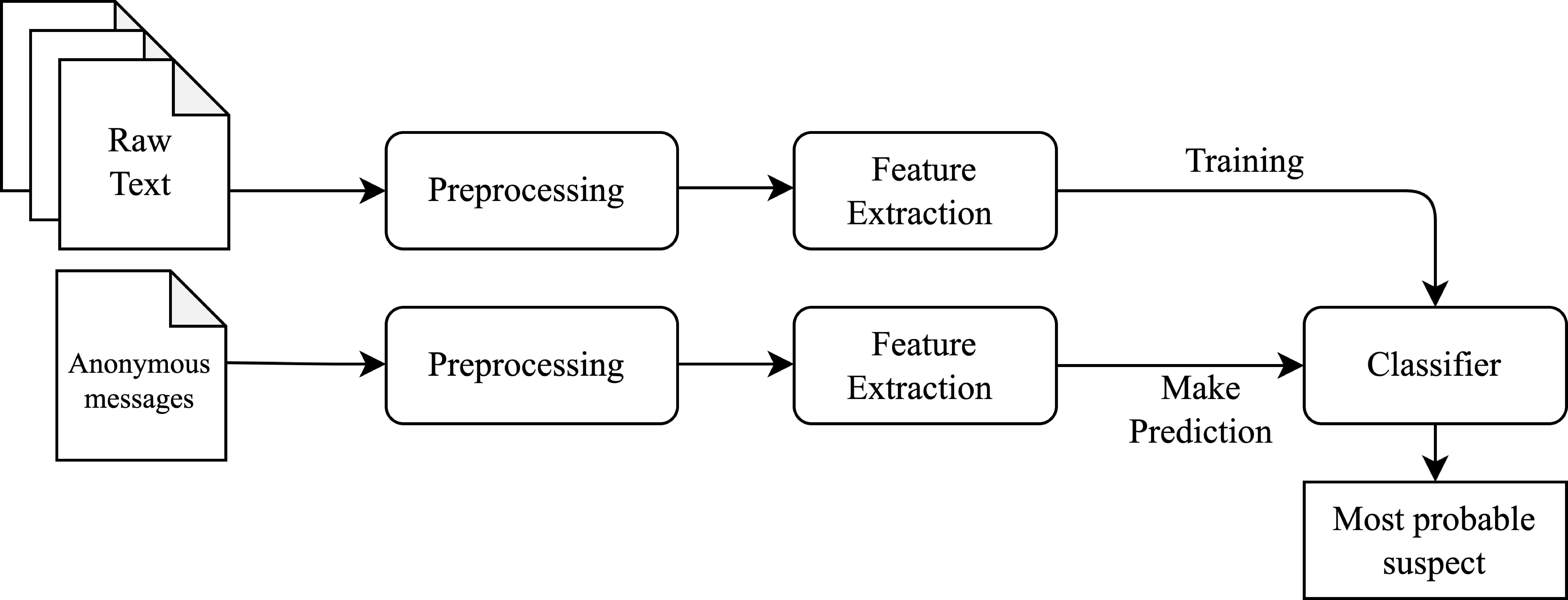}
\caption{Architecture of authorship attribution.} \label{fig1}
\end{figure}

Online predator attribution aims to establish the authorship of a harmful text typed by a specific predator from a set of known predators using his stylistic traits. To accomplish this goal, most of the author attribution systems follow the general structure illustrated in Figure \ref{fig1}. In the following subsections, a detailed description of each component will be provided.  

\subsection{Datasets}
Pender has stated in \cite{pendar2007toward} that there are two types of online text chat with sexually explicit content: consensual interaction between adults, and interaction between a sexual predator and what that individual believes to be a victim.  Focusing on the second type, he concluded that three types of interaction could happen: predators with underage victims, predators with law enforcement officers pretending to be a child, and predators with volunteers posing as underage teenagers. Accessing the first two predatory conversations is difficult mainly for privacy and legal issues. So, the only option available is the last one: using chat text of predators with volunteers posing as children.

One of the most used datasets that have adopted this approach is the \\
Perverted - Justice (PJ) chat logs. The data is available through\\ http://www.perverted-justice.com where it was operated by a group of volunteers. They post archives of pedophiles’ text chats with contributors posing as underage children. 

Also, the published dataset of PAN 2012 in \cite{inches2012overview} has been used repeatedly in the literature. Although it was mostly used for the task of identifying predators, the corpus is suitable for the attribution task as well. The authors of \cite{inches2012overview} had constructed the dataset from four different resources: Omegle\footnote{http://omegle.inportb.com/} to represent the adult consensually sexual conversations, IRClog\footnote{http://www.irclog.org/} and Krijn\footnote{http://krijnhoetmer.nl/irc-logs/} to represent regular online conversations, and the PJ chat logs mentioned previously. The first three resources were used to add a false negative set of text to the true positive data in PJ. In some cases, researchers opted to use data scraped from social media like Twitter. For example, the dataset used in \cite{imperial2021pedophiles} has raw tweets related to child sex trafficking and peddling that had been published within the area of the Philippines. They used hashtags that were reported to be commonly used by child sex traffickers as bookmarks for their tweets. This dataset (as is usually the case with similar this type of data) was not publicly published.

On the other hand, authors of \cite{neal2017surveying} have listed 13 publicly available data sets for the general author attribution task, none of which are related to pedophiles or sexual predators.

\subsection{Preprocessing}
Preprocessing is a necessary step to remove the noise and filter unrelated data to improve
the performance of the model and reduce the processing time. Depending on the acquired data, the preprocessing phase might include normalization and noise removal. Normalization is to unify the document format, as the raw messages are gathered in a semi-structured textual format such as XML or Jason file. This action grants that all the data is in the same structure. In \cite{sekeres2020advanced} the author was able to develop a method that can accomplish this step in an automatic/semi-automatic way.

Noise removal is needed to improve the performance of learning and reduce the training time.
At this stage, noise removal includes removing noisy conversion (at the sample level) by removing non-textual samples, conversations that include only one participant, and short messages and short greetings. The removed elements should not affect the learning process.

\subsection{Feature Extraction}

Feature extraction is identifying important features or attributes of the text to reduce the
initial set of raw data to a more manageable and meaningful set for processing. Table \ref{tab1}  lists the common features used in the online predators’ attribution systems.
 When extracting author stylistic features to attribute the text, two approaches are mainly used: The profile-based approach, which joins all samples of an author to extract a single feature vector, and the Instance-based approach which extracts a single feature vector per document. Choosing between these two models is mostly related to the problems of classification (discriminative, generative) and the kind of writing style (general style for each author or a separate style for each document) \cite{stamatatos2009survey}.
\subsubsection{A. Feature Extraction Approaches}
\begin{figure}
\includegraphics[width=\textwidth]{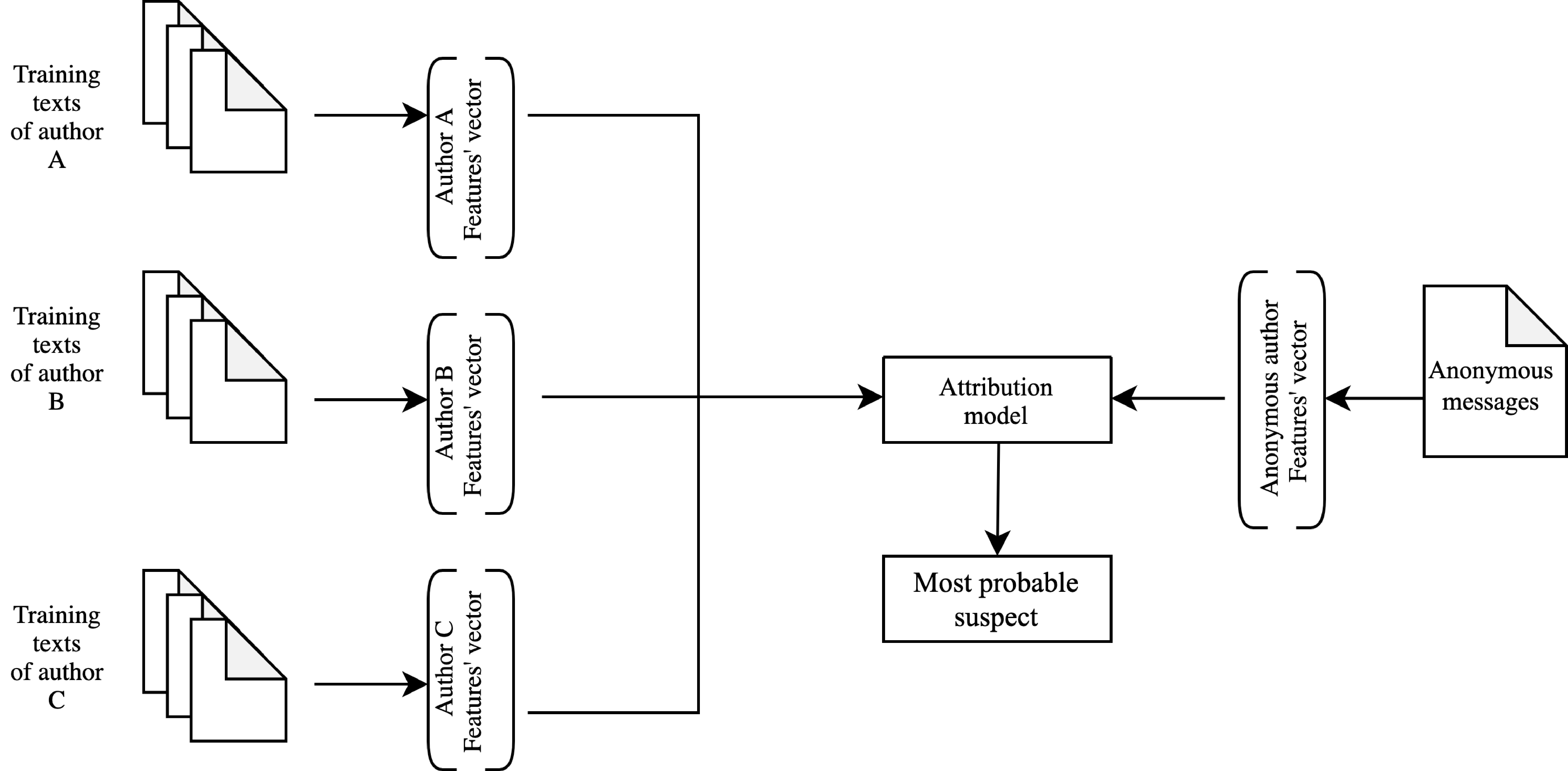}
\caption{Architecture of profile-based approach.} \label{fig2}
\end{figure}
In \cite{inches2013finding} profile-based authorship attribution, an author profile is constructed by obtaining several sample texts for each author, then extracting the relevant information from each sample and concatenating them in a single document or profile. An anonymous text is linked with the author whose profile is most similar to the anonymous text. Profile-based method excels in its simplicity and efficiency, scalability, and interpretability \cite{escalante2011weighted}.
\begin{figure}
\includegraphics[width=\textwidth]{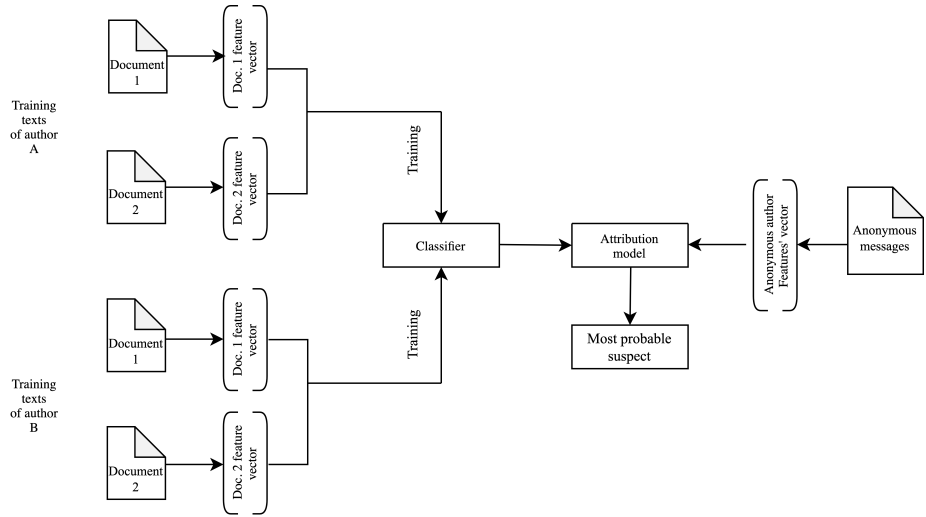}
\caption{Architecture of instance-based approaches.} \label{fig3}
\end{figure}
On the other hand, the instance-based approach produces a vector of features for each text sample in the training corpus and passes them through a classifier to develop an attribution model that can predict the author of an unseen text. So, to train the classifier multiple samples for each author are required. The majority of the work done in authorship attribution belongs to this category. The popularity of this approach is mostly related to its ability to utilize machine learning algorithms to handle noisy, high-dimensional and sparse data \cite{stamatatos2009survey}. Satmatatos in \cite{stamatatos2009survey} has provided an excellent compression between the two approaches.

It is worth mentioning that in some cases a Hybrid approach is adapted to combine the two to benefit from both of their characteristics. For example, van Halteren in \cite{halteren2007author} has adopted an approach to represent the training sample separately, similar to the instance-based approach. Then he averaged the represented vectors of each author to produce a single profile vector for each author, similar to the profile-based approach.

\subsubsection{B. Features}

Many features can be used to classify an author, but from a forensic perspective, there is the need to identify the correct features that can be included in the attribution process \cite{amuchi2012identifying}. The most used features for identifying predators' writing styles are lexical features. The reason is lexical features are language-independent and can capture subtle differences in style and contextual information \cite{neal2017surveying}. These features divide the text into character and word-based groups to capture stylistic traits at each level. Lexical features used include a bag of word approach (BOW), Term Frequency- Inverse Document Frequency (TF-IDF) word embeddings, character and word N-grams,  skip grams, Hashing Vectorizer, the length of the text, count/ratio of “emoticons”, count/ratio of profanity, and the number of pronouns. However, some lexical features (word-based) can produce a lot of noise if the text has misspellings or abbreviations \cite{rocha2016authorship}.

 Also, other used features in the literature are: syntactic (which captures style in the organization of sentences), structural (which illustrates how an author organizes the document), and domain-specific (which includes frequency of keywords or other content-specific information in the document).

Some feature representations are difficult to classify as they might include several feature categories. For example, authors of \cite{imperial2021pedophiles} have used topic modelling features to have a deeper understanding of the role played by pedophiles in social media platforms. In addition, \cite{belvisi2020forensic} authors have used idiosyncratic features to capture spelling and grammar mistakes and words that reflect social and cultural background. Also, the authors of \cite{iqbal2013unified} have used the writeprint of a suspect as a feature to uniquely identify him from a pool of other suspects. 

Authors of \cite{amuchi2012identifying,inches2013finding,iqbal2013unified,ishihara2013comparative} have used syntactic features in their approach to attribute suspects to the text. These features produce reliable results and improve the overall performance. Still, syntactic features depend heavily on the quality of tokenizers and parsing tools \cite{rocha2016authorship} and it doesn’t capture important information when used with very short text like Twitter messages \cite{belvisi2020forensic}. Moreover, when the document has a distinctive structure like emails and tweets, using structural features can help in capturing the suspect's style. But these features might show some sensitivity towards unbalanced data \cite{belvisi2020forensic}.

 Several studies have attempted to examine the performance of their models using different feature sets \cite{amuchi2012identifying,belvisi2020forensic,rocha2016authorship}. The results have shown that using a combination of features always gives the best performance, especially when lexical features (e.g. character n-grams) are used. In \cite{neal2017surveying} authors have stated that character n-grams and function words are the most reliable and efficient features for authorship attribution.
 
In some cases, a second noise removal step is needed after feature extraction to remove noisy features, like emoticons or small images, unintentionally misspelled words or terms not in the proper encoding like using multiple languages in the raw text or using encoding different from UTF \cite{amuchi2012identifying,ebrahimi2016automatic}.
\begin{table}
\caption{Features used in predator attribution.}\label{tab1}
\begin{tabular}{|l|l|l|}
\hline
\bf{Features}& \bf{Description} &\bf{Example}\\
\hline\hline
\bf{Lexical}&\multirow{4}{5cm}{Divide the document into groups of characters and words to capture stylistic attributes.} & - No. of characters \\
\cite{amuchi2012identifying},\cite{belvisi2020forensic},\cite{imperial2021pedophiles},\cite{iqbal2013unified},\cite{ishihara2013comparative},\cite{misra2019authorship}&  &- No. of words\\
&&- No. of digits\\
&&- Average word length\\
\hline\hline
\bf{Syntactic}&\multirow{3}{5cm}{Capture style in the organization of sentences.} & - Punctuation frequency  \\
\cite{amuchi2012identifying},\cite{inches2013finding},\cite{iqbal2013unified},\cite{ishihara2013comparative}&& - Function word frequency\\
&&- Part-of-speech (POS)\\
\hline\hline
\bf{Structural}&\multirow{3}{5cm}{Illustrate how an author organizes the document}.&- No. of words per paragraph\\
\cite{belvisi2020forensic}, \cite{iqbal2013unified} &&- Font size and colour\\
&&- Paragraph length\\

\hline\hline
\bf{Domain-Specific}&\multirow{2}{5cm}{Refer to content-specific features in the document.}&- Content-specific terms\\
\cite{amuchi2012identifying}, \cite{iqbal2013unified} &&- Language-specific words\\

\hline\hline
\bf{Additional Features:} && \\
Topic Modelling&\multirow{1}{5cm}{Treat document as a collection of topics,  where certain topics would have a higher probability to have specific terms.}&- Word distributions\\
\cite{imperial2021pedophiles}&&\\
&&\\
&&\\
\hline
Idiosyncratic Features&\multirow{3}{5cm}{Capture elements that are unique to the author.} & - Set of misspelled words\\
\cite{belvisi2020forensic}&&- Abbreviations\\
&&- Use of emojis\\
\hline
Writeprint&\multirow{1}{5cm}{Extract individualized features per author.}&- Lexical\\
\cite{iqbal2013unified}&&-Syntactic\\
&&-Structural\\

\hline\hline
\end{tabular}
\end{table}

\subsection{ Classification}
After extracting features from the corpus, classification is used to learn predatory patterns and attribute each conversation to its author. In the classification step, train and test features are compared to measure the distance between them. When approaching the online predator attribution problem, most of the work used either machine learning-based or distance-based methods to classify the feature space. 

Machine learning algorithms can be used for classification and clustering. Supervised machine learning algorithms are adopted when the class labels are known and work best with multidimensional and sparse features. This approach is widely used in the stylometry task, for example, all the participants in the PAN 2013 \cite{rangel2013overview} and most of the participants in PAN 2012 \cite{inches2012overview} competitions have used this approach in their models. The most popular choice among researchers in this category is the Support Vector Machine algorithm (SVM). The reason is its ability to deal with large and sparse datasets \cite{rocha2016authorship}, and it was the best performing algorithm in \cite{amuchi2012identifying} with 85.80\% accuracy. Some researchers used Neural networks for classification as they handle diverse feature sets very well and are suitable for learning models for nonlinear problems \cite{neal2017surveying}. For instance, \cite{misra2019authorship} have proposed two models: AA-CNN for author attribution, and AA- CNN- PC for predator classifications and attribution that use Convolution Neural Network CNN. They were able to report up to 92\% accuracy for the closed-set experiment, and 63\% accuracy for the open-set experiment.  On the other hand, \cite{imperial2021pedophiles} and \cite{iqbal2013unified} chose a clustering approach (unsupervised learning).  Clustering generates clusters or groups of unlabeled data points where each group has similar characteristics or features. This algorithm is mainly used to find stylistic similarities or to reduce the search space for authorship attribution problems \cite{neal2017surveying}.

Distance-based methods measure the similarity (distance) between the vocabulary used in two texts to determine if the same person had written them. The smaller the distance between documents, the more similar they are. Some of the commonly used distance metrics: are Delta, which is the most popular, Chi-Square Distance in, Kullback-Leibler Divergence, and Cosine similarity.

In \cite{amuchi2012identifying} authors have compared two distance methods Kullback-Leibler Divergence (KLD) and Chi-squared. They have found that in closed set attribution with a small set of authors (20 authors) KLD has slightly better performance (92\%) Chi-squared (90\% accuracy). Also, when tested on the open set attribution, KLD had the highest performance with 63\% accuracy. In \cite{belvisi2020forensic}, authors have followed a unique approach by examining the performance of each feature individually. In their work, the authors have compared three distance-based methods (Cosine, Euclidean and Manhattan.) to assess the performance of three features (lexical, structural and idiosyncratic) independently. They found that the Cosine distance is superior to the other two distances. And the best performance feature was the idiosyncratic features with 98.5\% accuracy. In \cite{iqbal2013unified} authors proposed a method to identify suspects with small training samples, in which they clustered the anonymous emails by stylometric features and then extracted the frequent stylometric patterns and writeprints from each cluster, and finally identified the suspect by comparing the writeprints of the training samples and the anonymous email clusters. With their approach, they were able to achieve 80\% accuracy with author sets equal to four users.

In some cases, the performance of distance-based methods is not the best when compared with machine learning-based methods. For instance, authors of \cite{amuchi2012identifying} have found that SVM and Naive Bayes (with 85\% and 84\% accuracy respectively)  outperformed the Chi-Square distance algorithm (60.90\% accuracy).

However, other approaches have been mentioned in the literature to attribute text to predators. For example, the work of \cite{ishihara2013comparative} has compared two probabilistic models (Multivariate kernel density and Gaussian mixture model – universal background model) to calculate the likelihood ratio in the same set of text.

\begin{table}
\caption{Summery of methods performance.}\label{tab1}
\begin{tabular}{|l|l|l|l|l|l|}
\hline
\bf{Ref. }&  \bf{Data }& \bf{No. of }& \bf{Attribution}  & \bf{Classification}  & \bf{Accuracy}\\
&\bf{type}&\bf{authors}&\bf{approach}&\bf{method}&\\
\hline\hline
\cite{amuchi2012identifying} & Chat logs & 10  & Profile&-Naive Bayes,& Best accuracy:\\
&&(closed-set)&based&-SVM&SVM: 85.80\% \\
&&&&-Bayesian Regression&\\
&&&&-Markov Chain&\\
&&&&-Chi-Square&\\
\hline
\cite{belvisi2020forensic}& Twitter& 40& Profile&-Cosine distance& Idiosyncratic: 98.5\%\\
&posts&(closed-set)&based &-Euclidean distance&Lexical: 96.0\%\\
&&&&-Manhattan distance&Structural: 96.0\%\\
\hline
\cite{inches2013finding}& Chat logs & 20 & Hybrid &-Chi-squared,&Best accuracy\\
&&(closed-set)&&-Kullback-Leibler& KLD: 92\% (closed-set)\\
&&up to 19046&&Divergence& KLD: 63\% (open-set)\\
&&(open-set)&&&\\
\hline
\cite{iqbal2013unified}& Emails& from 4 to 20 & Profile &-Frequent pattern-& 80\% - 40\%\\
&&(closed-sets)&based&based write-print &\\
&&&&mining with clustering&\\
\hline
\cite{ishihara2013comparative}& Chat logs & 115& Instance& Likelihood ratio (LR):& Best Log LR cost:\\
&&(closed-set)&based&-MVKD&GMM-UBM: 0.268\\
&&&&-GMM-UBM&\\
\hline
\cite{misra2019authorship}& Chat logs & 10 and 50& Instance&-CNN & AA-CNN: 55.70\%\\
&&(closed-sets)&based&&AA-CNN-PC: 54.90\%\\
\hline\hline
\end{tabular}
\end{table}

\section{Performance Analysis}
The majority of reviewed articles related to identifying and attributing sexual predators have used computational performance matrices without utilizing user studies or human evaluation methodologies \cite{razi2021human}. Some of the performance computational measures that have been used are accuracy, recall, precision and F1-score.

Researchers agreed that using accuracy as the only measure of performance is not recommended, as it tends to be biased with unbalanced datasets. Also, in our case where the goal is to attribute an anonymous text to a pedophile, precision (percentage of the actual suspects) is more desirable than recall (percentage of all the possible suspects). Since law officers’ target is to catch the actual online predator rather than widen the pool of suspects.

\section{Existing Author Attribution Tools}
According to the authors of \cite{razi2021human}, few studies proposed tools to mitigate the risk of online sexual predators, like the work of \cite{macfarlane2009agent} and \cite{penna2010framework} to detect and filter sexual grooming content online, and the work of \cite{silva2014data} and \cite{wang2012data} which focus on removing sex trafficking materials.

On the other hand, several tools have been developed to assess experts and non-experts in the general task of authorship attribution. Note that these tools focus only on the closed-form attribution problem with long text samples, hence does not fit the forensic or legal field criteria and was not built for that purpose. Some of these tools:
\begin{description}
\item[JGAAP]\footnote{https://github.com/evllabs/JGAAP/blob/master/docs} \cite{juola2006prototype} (Java Graphical Authorship Attribution Program). Researchers developed the first version of this tool in EVL labs at Duquesne University in 2013. It is a Java-based application that takes a text file as input, and then the user can choose between different preprocessing techniques, different feature selections, and a wide variety of classification algorithms. This tool supports several languages, and the produced output is very clear and easy to understand.

\item[Authorship attribution software.]\footnote{http://www.aicbt.com/authorship-attribution/online-software/} It was developed in 2013 at the AICBT, a research facility in Vancouver. This tool provides very few options when compared to the former one. Although it is web-based and easily accessible, it is only able to compare two users at a time. Overall, it is easy to use especially for non-experts users however the result is not conclusive.

\item[Stylo.]\footnote{https://github.com/computationalstylistics/stylo} \cite{eder2016stylometry} This R package was first built in 2016 by Maciej Eder, Jan Rybicki, Mike Kestemont, and Steffen Pielstroem. As the first tool, this package supports several languages and offers customized attribution steps. It allows the users to load their own annotated corpora to use it as a part of its processing pipeline. However, it doesn’t replace other domain-specific NLP tools, but it can accommodate their output in its pipeline. One of its big advantages is presenting the results in cluster visualization.

\end{description}

\section{Open Research Problems}
We established in the previous sections how the task of attributing Pedophiles to their texts is challenging due to several reasons (size of the training sample, number of suspects, length of the text, specific requirements of the domain). As a consequence, most of the attribution approaches in the literature focused on these aspects. Yet, to advance this domain, many research directions must be investigated by the research community.

As mentioned earlier, the most used datasets to identify online predators are the PJ and PAN-2012. Both datasets had adults posing as children to lure predators, hence no real victims were involved. The reason is acquiring a real dataset for underage victims is difficult due to the involvement of law agencies \cite{inches2012overview}. Although, this strategy may work when the goal is to detect the behaviour or identity of predators, yet, it would fail to capture adolescents’ behaviour \cite{razi2021human}. Considering real victims’ behaviour and interactions can help in the development of methods that prevent those underage teens from being victimized \cite{doring2014consensual,razi2020let}. Consequently, there is a need to construct real datasets with real underage victims to capture this type of information. 

Also, using data from social media and chat logs has always posed a challenge in the field of digital forensics. For instance, in author attribution, having a large number of suspects and few samples for each of them is common, which means that there is a rarity in the training text material \cite{stamatatos2009survey}, and we might even have only one message for certain suspects \cite{rocha2016authorship}. For that reason, new research directions should consider the use of multi-modal data and diverse feature sets. In \cite{wang2012data} authors have suggested using a multi-modal data approach, where images and meta-data can be used with textual data to better understand the complexity of the real world. Also, the author of \cite{dan2018identifying} had included behavioural features and lexical features to identify predators, and authors of \cite{kanakaris2021shall} had used graph theory to represent the author’s network in addition to textual-based features.

Usually, the accuracy of an authorship attribution system depends on the number of possible authors, the length of the texts, and the size of the training set \cite{luyckx2011effect,stamatatos2009survey}. However, if we consider the setting of attributing pedophiles, we need additional criteria to meet court standards. The reason is the cost of the wrong prediction in these systems is high which could lead to legal ramifications and wrongful convictions \cite{rocha2016authorship}. Therefore, future work should adapt not only a method to evaluate the significance of certain features (character or lexical) for specific authors but also a high-level abstract description of the author’s style \cite{stamatatos2009survey}. 

Moreover, when evaluating the performance of author attribution approaches for sexual predators, researchers mostly used computational methodologies. Evaluating models using computational matrices like recall or percussion only reflects the computational performance of the model, and does not assess if the model is suitable for a real-life scenario \cite{razi2021human}. Hence, future research should consider covering all aspects of the system when evaluating these models by incorporating surveys and user studies (i.e. human-centred evaluation) with computational evaluation  \cite{mohseni2021multidisciplinary}.

\section{Conclusions and Future Work}
As the number of social media users increases rapidly, the number of predators on these platforms multiplies as well. Protecting vulnerable users, especially children, by providing a safe environment to communicate is important. However, having an effective forensic solution is challenging and requires serious attention from researchers in this area. One of the main problems and most difficult in the field is authorship attribution. In this survey, we reviewed several studies on computational techniques for online pedophile attribution. We defined the problem from different perspectives and listed the main parts of an attribution system, in addition to commonly used performance measures. We reviewed a few of the existing attribution tools and their limitations. Then, we provided future research directions for the open research problems in the field.

Through our review of the literature, all the articles we surveyed have provided an algorithmic solution without disclosing the possibility of implementing their methods into a real-world system or application, nor provided a discussion on the type of users we need to consider for these systems. According to \cite{arrieta2020explainable}, to trust the outcome of these systems we need to provide explainability of the AI system’s functioning or decisions to be understandable by respected stakeholders. Thus, we would like to extend this study to build and examine the performance of a specialized pedophile attribution system when explainable AI techniques are used.

\bibliographystyle{splncs04}
\bibliography{Survey}

\end{document}